\documentclass[letterpaper, 10 pt, conference]{ieeeconf}  

\IEEEoverridecommandlockouts                              

\usepackage{graphicx}
\usepackage{cite}
\usepackage{amsmath,amssymb,amsfonts} 
\usepackage{algorithmic}          
\usepackage{float}                
\usepackage{booktabs}             
\usepackage{multirow}             
\usepackage{textcomp}             
\usepackage{comment}             
\usepackage[table]{xcolor}       
\usepackage{colortbl}             
\usepackage{censor}

\makeatletter
\let\NAT@parse\undefined

\definecolor{bestblue}{RGB}{220,235,255}
\def\BibTeX{{\rm B\kern-.05em{\sc i\kern-.025em b}\kern-.08em
    T\kern-.1667em\lower.7ex\hbox{E}\kern-.125emX}}

\newcommand{\pmstd}[1]{\,{\scriptsize$\pm$\,#1}}
\newcommand{\mspm}[2]{#1\pmstd{#2}}         
\newcommand{\bmspm}[2]{\begingroup\setlength{\fboxsep}{0.6pt}\colorbox{bestblue}{\strut\textbf{#1}\pmstd{#2}}\endgroup} 
\newcommand{\mspmNA}{--}

\usepackage[colorlinks=true,citecolor=blue]{hyperref} 
\title{\LARGE \bf
MTA-RL: Robust Urban Driving via Multi-modal Transformer-based 3D Affordances and Reinforcement Learning
}

\author{Guangli Chen\textsuperscript{*}, Dianzhao Li\textsuperscript{*}, Wenjian Zhong, Bangquan Xie\textsuperscript{\textdagger} and Ostap Okhrin
\thanks{\normalfont\textsuperscript{*} These authors contributed equally to this work.}
\thanks{\normalfont\textsuperscript{\textdagger} Corresponding author: Bangquan Xie}
\thanks{Guangli Chen is with Dongguan Key Laboratory of Intelligent Equipment and Smart Industry, School of Advanced Engineering,  Great Bay University, Dongguan, China; Chair of Applied Statistics, Technische Universität Dresden, Dresden, Germany.
        {\tt\small (guangli.chen@mailbox.tu-dresden.de)}}%
\thanks{Dianzhao Li and Ostap Okhrin are with  Chair of Applied Statistics, Technische Universität Dresden, Dresden, Germany; 
Center for Scalable Data Analytics and
Artificial Intelligence (ScaDS.AI) Dresden/Leipzig, Dresden, Germany.
        {\tt\small (dianzhao.li@tu-dresden.de, ostap.okhrin@tu-dresden.de)}}%
\thanks{Wenjian Zhong is with College of Automation, Guangdong University of Technology, Guangzhou, China.
        {\tt\small (wenjianzhong973008@163.com)}}%
\thanks{Bangquan Xie is with Dongguan Key Laboratory of Intelligent Equipment and Smart Industry, School of Advanced Engineering,  Great Bay University, Dongguan, China.
        {\tt\small (bangquanxie@gbu.edu.cn)}}%
}

\begin{document}

\maketitle
\thispagestyle{empty}
\pagestyle{empty}

\begin{abstract}

Robust urban autonomous driving requires reliable 3D scene understanding and stable decision-making under dense interactions. However, existing end-to-end models lack interpretability, while modular pipelines suffer from error propagation across brittle interfaces. This paper proposes \emph{MTA-RL}, the first framework that bridges perception and control through \emph{M}ulti-modal \emph{T}ransformer-based 3D \emph{A}ffordances and \emph{R}einforcement \emph{L}earning (RL). Unlike previous fusion models that directly regress actions, RGB images and LiDAR point clouds are fused using a transformer architecture to predict explicit, geometry-aware affordance representations. These structured representations serve as a compact observation space, enabling the RL policy to operate purely on predicted driving semantics, which significantly improves sample efficiency and stability. Extensive evaluations in CARLA Town01–03 across varying densities (20–60 background vehicles) show that \emph{MTA-RL} consistently outperforms state-of-the-art baselines. Trained solely on Town03, our method demonstrates superior zero-shot generalization in unseen towns, achieving up to a 9.0\% increase in Route Completion, an 11.0\% increase in Total Distance, and an 83.7\% improvement in Distance Per Violation. Furthermore, ablation studies confirm that our multi-modal fusion and reward shaping are critical, significantly outperforming image-only and unshaped variants, demonstrating the effectiveness of \emph{MTA-RL} for robust urban autonomous driving.
\end{abstract}

\section{INTRODUCTION}

Urban autonomous driving requires vehicles to operate under partial observability and uncertainty while simultaneously solving tightly coupled tasks such as navigation, trajectory planning, dynamic obstacle avoidance, traffic rule compliance, and safe interaction with vulnerable road users\cite{wu2024recent,sallab2016end,b5,zhou2019development,li2024interactive,li2025ethics,zhao2024autonomous}. To address this challenge, two dominant paradigms have emerged: \emph{modular pipelines} and \emph{end-to-end learning}. While modular approaches \cite{van2018autonomous,li2025autonomous} offer interpretability, they heavily rely on hand-crafted priors, and upstream errors can propagate downstream, degrading robustness in highly dynamic scenarios \cite{mcallister2017concrete}. Alternatively, end-to-end learning aims to overcome these limitations by directly mapping sensory inputs to control commands \cite{tampuu2020survey,chen2024end}. Within this paradigm, imitation learning (IL) \cite{le2022survey} is data-efficient but suffers from compounding errors caused by distribution shifts in long-tail scenarios. Unlike IL, RL\cite{sutton1998reinforcement} is capable of surpassing expert demonstrations, but it often exhibits poor sample efficiency and limited interpretability when processing high-dimensional raw observations. Moreover, by tightly coupling perception and decision-making, pure end-to-end RL lacks explicit spatial reasoning, making it difficult to interpret failures or guarantee safety in novel environments.

To bridge the gap between modular pipelines and end-to-end learning, \emph{direct perception} has been proposed as an intermediate paradigm \cite{chen2015deepdriving}. Direct perception learns compact, task-relevant intermediate representations from raw sensory inputs, commonly referred to as \emph{affordances}. These affordances encode essential driving information such as ego-vehicle states, traffic rules, and interactions with surrounding agents, enabling decision-making in a structured and semantically meaningful state space. By decoupling perception from control, affordance-based systems improve interpretability and robustness, and allow downstream control to be implemented using either rule-based \cite{sauer2018conditional} or learning-based policies \cite{ahmed2021policy}.

Despite these advantages, most existing affordance learning approaches rely solely on monocular camera images, restricting the learned representations to the 2D image plane and limiting their geometric expressiveness. This weak 3D geometric reasoning leads to unstable policies in dense urban traffic. Overcoming this requires multi-modal fusion. While standard CNN-based fusion methods often suffer from inherent viewpoint misalignment, leading to depth and feature blurring across modal views, and incur significant geometric information loss when forcing sparse 3D points into grid-structured convolutions\cite{vora2020pointpainting,liang2018deep}, recent advances in transformer-based architectures \cite{vaswani2017attention,dosovitskiy2020vit} have demonstrated strong capabilities in multi-modal fusion, enabling effective integration of RGB images and LiDAR point clouds for richer 3D scene understanding \cite{vora2020pointpainting,natan2022end,sobh2018end}. However, how to leverage such multi-modal representations to support safe and efficient decision-making remains an open challenge.  

\begin{figure*}
    \centering
    \includegraphics[width=0.8\textwidth]{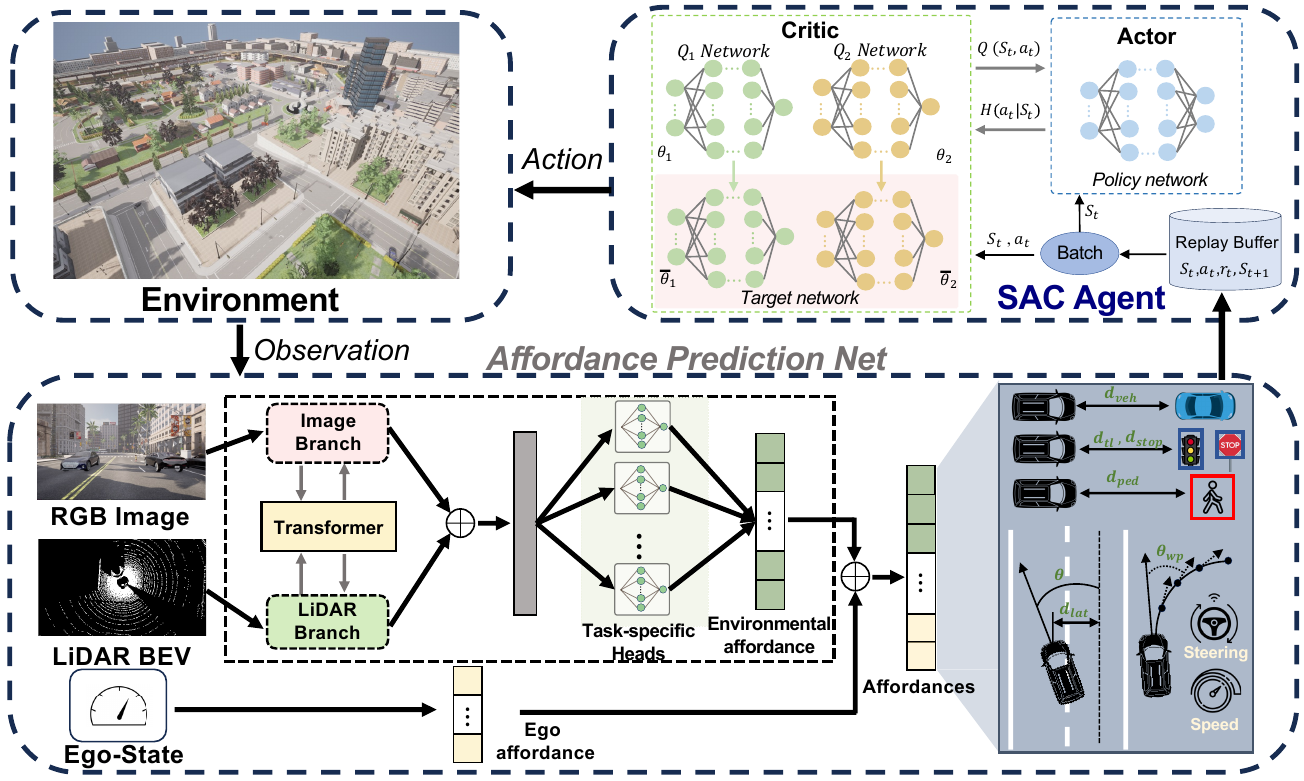}
    \caption{Overview of the MTA-RL framework. A transformer-based affordance prediction network fuses RGB camera images and LiDAR BEV inputs to produce compact, geometry-aware affordance representations. These structured affordances serve as input to an RL controller, enabling stable decision-making.}
    \label{fig:overall}
\end{figure*}

Motivated by these observations, we propose \emph{MTA-RL}, the first unified framework that couples multi-modal 3D Bird’s-Eye-View (BEV) affordances with RL for urban driving. \emph{MTA-RL} employs a transformer-based fusion network \cite{toromanoff2020end} to jointly process RGB images and LiDAR point clouds in BEV, producing compact, geometry-aware, and safety-critical 3D affordances. These learned affordances serve as the structured state representation for a dedicated RL controller, as illustrated in Fig.~\ref{fig:overall}. Operating in this semantically grounded space improves sample efficiency, interpretability, and robustness, while facilitating the integration of traffic rules and explicit safety constraints through reward shaping and termination conditions \cite{sauer2018conditional,zeng2019end}. Our contributions are summarized as follows:
\begin{itemize}
\item \textbf{3D BEV affordance}: We propose a transformer-based multi-modal architecture with explicit distance-to-constraint heads to predict compact, geometry-aware affordances in BEV space.
\item \textbf{RL operating purely on predicted affordances}: The learned affordances provide a structured observation space, enabling stable RL policy learning in complex urban environments.
\item \textbf{Robust urban driving performance}: Extensive evaluations in CARLA demonstrate that \emph{MTA-RL} consistently outperforms state-of-the-art baselines in both affordance prediction accuracy and driving performance across diverse towns and traffic densities.
\end{itemize}

\section{RELATED WORKS}
\label{sec:related}

\subsection{Reliable Driving Systems}

Urban driving is fundamentally constrained by the limitations of individual sensing modalities: camera-based perception provides rich semantic information but suffers from inaccurate depth estimation and sensitivity to adverse illumination, whereas LiDAR offers precise geometric measurements but lacks semantic and texture cues \cite{xiao2020multimodal}. Consequently, multi-modal sensor fusion has become a critical component of modern AD systems. Early fusion methods relied on explicit geometric alignment, such as projecting LiDAR points into image space \cite{vora2020pointpainting} or combining features using continuous convolutions \cite{liang2018deep}. Although effective under ideal calibration, these approaches are sensitive to alignment errors and typically aggregate information locally, limiting their ability to capture global cross-modal interactions. More recently, transformer-based architectures have significantly advanced multi-modal fusion by leveraging attention mechanisms to model long-range dependencies across modalities. Notably, TransFuser \cite{chitta2022transfuser} employs self-attention to softly associate image and LiDAR features, substantially outperforming rigid geometric fusion approaches. However, most transformer-based fusion models are trained using IL and directly regress waypoints or control commands. Such formulations remain susceptible to causal confusion and distribution shift \cite{codevilla2019exploring}, often resulting in fragile behavior in rare or safety-critical situations. To achieve robust performance in such critical scenarios, our framework explicitly decouples perception from decision-making to empower a subsequent RL policy.

\subsection{Affordance Learning and Intermediate Representations}
\label{subsec2}

Early affordance learning methods, such as DeepDriving \cite{chen2015deepdriving} and CAL \cite{sauer2018conditional}, successfully extract task-relevant indicators (e.g., lane distances and traffic light states) to support downstream decision-making. However, these methods primarily rely on hand-crafted, rule-based controllers, which struggle to handle the complex, dynamic interactions typical of dense urban environments. To address this, more recent approaches like \cite{ahmed2021policy} and \cite{toromanoff2020end} replace rule-based systems with RL policies. Despite achieving improved adaptability, these methods dominantly rely on single-modal camera inputs, restricting their representations to the 2D image plane. The inherent lack of explicit 3D geometric reasoning severely limits their ability to accurately model multi-agent interactions. In contrast, we construct multi-modal affordances directly in the 3D BEV space to restore the geometric expressiveness required for robust downstream control.

\subsection{Reinforcement Learning for Autonomous Driving}
\label{subsec3}

While RL has shown promise for autonomous driving \cite{wu2024recent,kiran2021deep}, ensuring safety and sample efficiency in complex urban environments remains a fundamental challenge. To address safety, a recent work\cite{11048675} augments the Soft Actor-Critic (SAC) algorithm \cite{haarnoja2018soft} with model predictive control to enforce hard constraints; however, such methods are typically evaluated in structured expressway scenarios and struggle to scale to dense, unsignalized urban intersections. For urban navigation, Roach \cite{zhang2021roach} employs a PPO policy, but it heavily relies on privileged simulator states, limiting its practical interpretability. More recently, world-model-based RL methods like Think2Drive\cite{li2024think2drive} improve efficiency but lack the explicit state representations required to enforce strict traffic rules and geometric safety boundaries. Unlike these implicit or privileged baselines, \emph{MTA-RL} operates purely on the predicted explicit affordances. Combined with dense reward shaping, this ensures strict semantic and geometric compliance in highly interactive urban traffic.

\section{METHODOLOGY}
\label{sec:method}

This section presents \emph{MTA-RL}, a unified framework that couples multi-modal 3D affordance learning with RL for urban autonomous driving. \emph{MTA-RL} first fuses heterogeneous sensor inputs to predict structured, safety-critical affordances in a geometry-aware BEV representation. These affordances then serve as the state space for an RL agent augmented with an explicit, rule-aware safety wrapper that enforces traffic regulations and interaction safety.

\subsection{Transformer-Based Multimodal Affordance Prediction}
\label{sec:affordance_prediction}

To enable robust decision-making, we adopt a transformer-based fusion architecture \cite{chitta2022transfuser} to integrate RGB and LiDAR inputs. These fused multi-modal features are subsequently utilized to predict structured, geometry-aware driving affordances.

\subsubsection{Multimodal Sensor Fusion} 

The proposed framework leverages two complementary sensing modalities: RGB camera images and LiDAR point clouds. For visual perception, three forward-facing RGB cameras (left $60^\circ$, center, right $60^\circ$) are employed, each with a horizontal field of view (FOV) of $120^\circ$. After distortion correction and cropping, the images are concatenated into a composite image of resolution $704 \times 160$, covering an effective FOV of $132^\circ$. This representation captures rich semantic cues such as traffic lights, pedestrians, and surrounding vehicles, and is processed using a ResNet-34 backbone \cite{he2016deep} to extract multi-scale image features. For geometric perception, LiDAR point clouds are projected onto a $32\,\mathrm{m} \times 32\,\mathrm{m}$ BEV grid centered on the ego vehicle, with a spatial resolution of $256 \times 256$ ($0.125\,\mathrm{m}$ per cell). Points are discretized into two height-based channels corresponding to ground and obstacle returns. Additionally, the 2D navigation goal is rasterized and embedded as a third channel, forming a three-channel BEV representation. This design introduces early spatial alignment between geometry-aware perception and route guidance. Following fusion, the multi-scale feature maps are aggregated into a single latent vector $\mathbf{z}$ via global average pooling. The latent vector $\mathbf{z}$ encodes global scene context, cross-modal interactions, and navigation intent. It serves as a shared input to all affordance prediction heads (Section~\ref{sec:affordance}), enabling consistent perception outputs and providing a compact, semantically structured representation for downstream policy learning.

\subsubsection{Affordance Prediction}
\label{sec:affordance}

To extract task-relevant driving semantics from the shared latent representation $\mathbf{z}$, we employ a multi-task prediction architecture with task-specific heads. Each head is implemented as a lightweight multi-layer perceptron (MLP) and predicts a single driving affordance. All heads operate in parallel on $\mathbf{z}$, forming a one-to-many prediction structure that promotes efficient feature sharing and joint optimization across tasks. All affordances are predicted in the ego-centric BEV coordinate frame and are expressed as $\mathbf{a}_j = \mathrm{MLP}_j(\mathbf{z})$, where $\mathbf{a}_j$ denotes the prediction of the $j$-th driving affordance. Based on their functional roles, we categorize the predicted affordances into \emph{discrete} (categorical) and \emph{continuous} (regression) affordances.

\paragraph{Discrete affordances}

Discrete affordances encode high-level semantic constraints associated with traffic rules and interaction safety: \textbf{Red traffic light} and \textbf{Stop sign} indicate active regulatory constraints, and \textbf{Hazard stop} signals imminent collision risks from pedestrians or vehicles.

\paragraph{Continuous affordances}
Continuous affordances capture geometric and relational information required for precise longitudinal and lateral control: These explicitly include: \textbf{Lateral distance} and \textbf{relative angle}, representing the ego vehicle’s offset and heading deviation from the lane centerline; \textbf{Target vehicle distance}, measuring the longitudinal gap to the leading vehicle; as well as \textbf{Traffic light distance}, \textbf{Stop sign distance}, and \textbf{Pedestrian distance}, which provide the continuous longitudinal distances to the respective semantic entities ahead.

\subsection{Perception-Guided Reinforcement Learning Policy}

Based on the predicted driving affordances from the perception module, we formulate high-level driving decision-making as an RL problem. We adopt SAC as the policy learning algorithm due to its stability and sample efficiency in continuous control settings. The following section details the RL formulation and training setup.

\subsubsection{Action Space and Observation Space}

At each timestep $t$, the agent outputs a continuous action vector $a(t) = (a_{\text{steer}}(t), a_{\text{long}}(t)) \in [-1, 1]^2$. Here, $a_{\text{steer}}$ controls the lateral steering, and $a_{\text{long}}$ specifies the longitudinal command, mapping positive and negative values to throttle and braking, respectively.

The observation $o(t) = (o^{\text{aff}}(t)^\top, o^{\text{ego}}(t)^\top)^\top$concatenates the perception-derived affordances and the ego-vehicle state. The affordance component is defined as:
\begin{equation}
o^{\text{aff}}(t) = (d_{\text{lat}}, \theta, d_{\text{veh}}, d_{\text{tl}}, d_{\text{stop}}, d_{\text{ped}}, I_{\text{red}}, I_{\text{stop}}, I_{\text{hazard}})_t^\top,
\notag
\end{equation}
where $d_{\text{lat}}$ and $\theta$ denote the lateral offset and relative heading. The variables $d_{\text{veh}}, d_{\text{tl}}, d_{\text{stop}}$, and $d_{\text{ped}}$ measure longitudinal distances to the leading vehicle, traffic light, stop sign, and pedestrian, respectively. The binary indicators $I_{\text{red}}, I_{\text{stop}}$, and $I_{\text{hazard}}$ indicate the presence of an active red traffic light, a stop-sign constraint, and imminent collision hazards induced by vehicles or pedestrians, respectively.

The ego-state component captures the current speed $v(t)$, as well as the steering and longitudinal commands from the previous time step:
\begin{equation}
o^{\text{ego}}(t) = (v(t), a_{\text{steer}}(t-1), a_{\text{long}}(t-1))^\top.
\notag
\end{equation}
By explicitly separating semantics from vehicle dynamics, this decoupled formulation yields a compact representation that facilitates stable policy learning.

\subsubsection{Reward Function}

The reward function jointly optimizes multiple objectives to promote robust and safe driving behavior in complex urban environments. At each time step $t$, the overall reward $R(t)$ is defined as:
\begin{equation}
\begin{aligned}
R(t)
&=
R_{\text{lane}}(t)
+ R_{\text{speed}}(t)
+ R_{\text{smooth}}(t)\\
&\quad
+ R_{\text{prog}}(t) 
+ R_{\text{rule}}(t)
+ R_{\text{event}}(t).
\end{aligned}
\label{eq:reward_total}
\end{equation}
To achieve an optimal balance among these diverse driving objectives, the specific weighting coefficients for each component were determined through extensive empirical evaluations. These terms respectively regulate lane-keeping, speed tracking, control smoothness, route progress, traffic rule compliance, and safety-critical terminal events, as detailed below.

\paragraph{Lane Keeping}
The lane-keeping reward penalizes both lateral offset and misalignment through smooth, bounded shaping terms:
\begin{equation}
R_{\text{lane}}(t)
=
w_{\ell 1}\cdot \exp\!\left(-2.5\,\lvert d_{\text{lat}}(t)\rvert\right)
+ w_{\ell 2}\cdot \frac{1}{1 + 3\,\lvert \theta(t)\rvert},
\label{eq:reward_lane}
\notag
\end{equation}
where $w_{\ell1}$ and $w_{\ell2}$ are set to $1.0$ and $0.2$, respectively. 

\paragraph{Speed Tracking}
To promote efficiency and penalize overspeeding, we define:
\begin{equation}
\begin{aligned}
R_{\text{speed}}(t)
&=
w_{s1} \cdot\Biggl(
1-\min\!\Bigl(
1,\,
\frac{\lvert v(t)-v_{\text{des}}(t)\rvert}{v_{\max}}
\Bigr)
\Biggr) \\
&\quad
+ w_{s2} \cdot\,
\frac{\max\!\bigl(0,\, v(t)-v_{\max}\bigr)}{v_{\max}},
\end{aligned}
\label{eq:reward_speed}
\notag
\end{equation}
where $w_{s1}=0.65$ and $w_{s2}=-0.7$ are fixed scalar weights, $v(t)$ denotes the current ego-vehicle speed, $v_{\text{des}}(t)$ is the desired speed, and $v_{\max}$ is the maximum allowable speed.

\paragraph{Control Smoothness}

To discourage abrupt actions and oscillatory behavior while promoting comfortable driving, we define a control smoothness term based on the step-to-step change in the steering command $a_{\text{steer}}(t)$ and the variation of the heading error $\theta(t)$ relative to the lane centerline:

\begin{equation}
\begin{aligned}
R_{\text{smooth}}(t)
= {} & w_{sm1} \cdot\,|a_{\text{steer}}(t)-a_{\text{steer}}(t-1)| + \\
     & w_{sm2} \cdot\,|\theta(t)-\theta(t-1)|, 
\end{aligned}
\notag
\end{equation}
where $w_{sm1}=-0.7$ and $w_{sm2}=-0.5$.

\paragraph{Route progress reward}

The progress reward is:
\begin{equation}
R_{\text{prog}}(t)
=
w_{pr1} \cdot\Bigl(
\Delta k(t)
-
\mathbb{I}_{\text{free}}(t)\bigl(1-\mathbb{I}_{\text{move}}(t)\bigr)
\Bigr),
\label{eq:r_prog}
\notag
\end{equation}
where $w_{pr1}$ is set to $0.4$, $\Delta k(t)$ denotes the waypoint index increment. $\mathbb{I}_{\text{move}}(t)=\mathbb{I}\!\left[\Delta k(t)>0\right]$ is indicator function, showing whether the vehicle is making progress. $\mathbb{I}_{\text{free}}(t)$, on the other hand, shows if the stopping constraint is triggered.

\paragraph{Traffic Rule Compliance Shaping}

To enforce safe and anticipatory stopping under traffic regulations, including traffic lights, stop signs, and pedestrian
yielding, we adopt a unified deceleration shaping strategy.
Let $d(t)$ denote the longitudinal distance from the ego vehicle to the active constraint, the traffic rule compliance shaping reward is defined as:
\begin{equation}
\begin{aligned}
R_{\text{rule}}(t)
=&\ w_{r1} \cdot\,\eta\left(d(t)\right)\cdot \frac{v(t)}{v_{\max}}+ \\
&\  w_{r2} \cdot\mathbb{I}\!\left[d(t)<2 \wedge v(t)<v_{\text{stop}}\right]
\cdot \!\left(1-\frac{v(t)}{v_{\text{stop}}}\right),
\end{aligned}
\notag
\end{equation}
where $w_{r1}=-1.8$ and $w_{r2}=1.5$, $\eta(d(t))=1-\min(d(t), D_{\text{stop}})/D_{\text{stop}}$ is a distance-based decay function, $D_{\text{stop}}$ is the effective stopping distance threshold, $v(t)$ is the ego speed, $v_{\text{stop}}$ is a small stopping threshold, and $\mathbb{I}[\cdot]$ is the indicator function. This shaping term penalizes late braking when approaching
stopping constraints and provides a small bonus for coming to a full stop. 

\paragraph{Termination components}
In addition to dense shaping rewards, we define event-based terminal conditions $R_{\text{event}}(t)$ for safety-critical violations and task completion. If a rule violation occurs or if the vehicle remains inactive for 50 consecutive steps, a one-time terminal penalty is applied and the episode is terminated. These terminal events and their corresponding rewards are summarized in Table~\ref{tab:terminal_penalty}. In contrast, reaching the destination yields a positive one-time reward without terminating the episode; instead, a new route is replanned and execution continues.

\begin{table}
\caption{One-time event reward.}
\label{tab:terminal_penalty}
\centering
\setlength{\tabcolsep}{6pt}
\renewcommand{\arraystretch}{1.05}
\footnotesize
\begin{tabular}{l c}
\toprule
\textbf{Event} & \textbf{$R_{\text{event}}(t)$} \\
\midrule
Red-light violation / STOP violation & $-100$ \\
Collision with vehicles/pedestrians & $-100$ \\
Collision with others & $-50$ \\
Off-lane / off-road & $-50$ \\
No progress (50 consecutive steps) & $-50$ \\
Reach destination & $+100$ \\
\bottomrule
\end{tabular}
\end{table}

\section{EXPERIMENTS AND RESULTS}

\subsection{Experiment Setting}

\subsubsection{Affordance Prediction Network}

To train the affordance prediction network described in Section~\ref{sec:affordance_prediction}, we construct a large-scale dataset using an autopilot agent in the CARLA simulator \cite{Dosovitskiy17}. The dataset contains approximately $220$k synchronized frames of RGB images, LiDAR point clouds, and ground-truth affordance annotations from various weather conditions and towns in CARLA, split into training, validation, and test sets. Training is conducted on 8 NVIDIA A100 GPUs. We adopt a multi-task learning objective to jointly optimize heterogeneous affordances, combining cross-entropy (CE) loss for categorical outputs and mean squared error (MSE) loss for continuous targets. The overall loss is defined as

\begin{equation}
\mathcal{L}_{\text{total}} 
= \sum_{c \in \mathcal{C}} \lambda_c \, \mathcal{L}_{\text{CE}}(a_c, \hat{a}_c)
+ \sum_{r \in \mathcal{R}} \lambda_r \, \mathcal{L}_{\text{MSE}}(a_r, \hat{a}_r),
\end{equation}
where $\mathcal{C}$ and $\mathcal{R}$ denote the categorical and regression affordance sets.  $\mathcal{L}_{\text{CE}}$ and $\mathcal{L}_{\text{MSE}}$ are the respective losses, with the weighting coefficients $\lambda_c$, and $\lambda_r$ balancing their respective contributions of individual prediction tasks during joint training.

\subsubsection{RL Policy Training}

RL experiments are conducted in the CARLA simulator. We select Town03 as the training map due to its diverse urban layouts, including traffic lights, stop signs, roundabouts, and complex intersections. At the beginning of each episode, start and goal locations are randomly sampled from predefined spawn points, and a point-to-point route is generated using a global route planner. Background traffic is managed by the CARLA Traffic Manager, with $60$ surrounding vehicles enabled and automatic lane changes activated. To simulate interactions with vulnerable road users, pedestrian crossing events are randomly triggered during training. Episodes terminate solely upon safety infractions or reaching the maximum step limit. Policy learning is performed using the SAC algorithm implemented in Stable-Baselines3 \cite{stable-baselines3}. Network architectures and hyperparameters are summarized in Table~\ref{tab:sac_hparams}. Training is carried out on an NVIDIA RTX 4060 GPU; $2$ million environment steps take about $13$ hours.

\begin{table}
\caption{Default hyperparameters of SAC Training.}
\label{tab:sac_hparams}
\centering
\footnotesize
\setlength{\tabcolsep}{6pt}
\renewcommand{\arraystretch}{1.05}
\begin{tabular}{l c}
\toprule
\textbf{Hyperparameters} & \textbf{Value} \\
\midrule
Neural network structure & $2 \times [256,\ \mathrm{ReLU}]$ \\
Learning rate ($\alpha$) & $3 \times 10^{-4}$ \\
Replay buffer size & $4 \times 10^{5}$ \\
Learning starts & $5 \times 10^{4}$\\
Batch size & $256$ \\
Discount factor ($\gamma$) & $0.99$ \\
Soft update coefficient ($\tau$) & $0.005$ \\
\bottomrule
\end{tabular}
\end{table}

\subsection{Evaluation Results}

\subsubsection{Evaluation of Transformer-Based Multimodal Perception}

To evaluate the proposed transformer-based affordance prediction network, we compare it with three representative baselines: \emph{DeepDriving} \cite{chen2015deepdriving}, \emph{CAL} \cite{sauer2018conditional}, and the affordance-based method of \cite{ahmed2021policy} (\emph{Affordance-RL}). Intersection-over-union (IoU) is reported for categorical affordances, and mean absolute error (MAE) for regression targets. Results are summarized in Table~\ref{tab:affordance}. Our method consistently outperforms all baselines on categorical affordances, including red light, stop sign, and hazard-stop detection. For continuous predictions, it achieves the lowest MAE on target vehicle distance and is the only approach that reliably estimates distances to traffic lights, stop signs, and pedestrians. Although some baselines perform competitively on simple lane-relative geometry, they fail to capture higher-level urban semantics and interaction-critical cues. These results demonstrate that the proposed multi-modal transformer architecture provides a more accurate and semantically comprehensive affordance representation for urban driving.

\begin{table}
    \centering
    \caption{Affordance prediction results on the test set. Best performance for each task is highlighted in bold.}
    \label{tab:affordance}
    \small
    \setlength{\tabcolsep}{4pt}
    \renewcommand{\arraystretch}{1.1}
    \resizebox{\columnwidth}{!}{%
    \begin{tabular}{lcccc}
    \toprule
    \textbf{Affordance} & \textbf{DeepDriving} & \textbf{CAL} & \textbf{Affordance-RL} & \textbf{MTA-RL(Ours)} \\
    \midrule
    \addlinespace[0.25em]
    \multicolumn{5}{@{}l@{}}{\textbf{Categorical} (IoU $\uparrow$)}\\
    \addlinespace[0.15em]
    Red traffic light & -- & 92.95\% & 98.97\% & \textbf{99.13\%}\\
    Stop sign & -- & -- & -- & \textbf{92.25\%}\\
    Hazard stop & -- & 87.36\% & 97.45\% & \textbf{98.76\%}\\
    \addlinespace[0.15em]
    \midrule
    \addlinespace[0.2em]
    \multicolumn{5}{@{}l@{}}{\textbf{Continuous} (MAE $\downarrow$)}\\
    \addlinespace[0.1em]
    Lateral distance & 0.309 & 0.085 & \textbf{0.030} & 0.124\\
    Relative angle & 0.032 & \textbf{0.002} & 0.083 & 0.019\\
    Target vehicle distance & 4.749 & 0.033 & 0.022 & \textbf{0.011}\\
    Traffic light distance & -- & -- & -- & \textbf{0.312}\\
    Stop sign distance & -- & -- & -- & \textbf{0.264}\\
    Pedestrian distance & -- & -- & -- & \textbf{0.236}\\
    \bottomrule
    \end{tabular}%
    }
\end{table}

\subsubsection{Evaluation of the driving performance}

\begin{table*}[t!] 
\centering
\footnotesize
\setlength{\tabcolsep}{3pt}
\renewcommand{\arraystretch}{1.05}

\caption{Evaluation results in dense traffic (60 BVs). Best results are highlighted in light blue.}
\label{tab:generalization_60}
\begin{tabular}{c l c c c c c c}
\toprule
Town & Model & AS$\uparrow$ & RC$\uparrow$ & TD$\uparrow$ & CR$\downarrow$ & CS$\downarrow$ & DPV$\uparrow$ \\
\midrule
\multirow{4}{*}{Town 1}
& VLM-RL       & \mspmNA & \mspm{0.630}{0.094} & \mspm{3340.4}{586.6} & \mspm{0.567}{0.115} & \mspm{16.34}{6.74} & \mspmNA \\
& Think2Drive  & \mspm{10.18}{1.23}  & \mspm{0.124}{0.001} & \mspm{709.1}{5.4}   & \bmspm{0.000}{0.000} & \bmspm{0.00}{0.00} & \mspm{70.9}{0.5} \\
& Roach        & \mspm{13.03}{1.44}  & \mspm{0.599}{0.144} & \mspm{3075.4}{863.4} & \mspm{0.033}{0.058} & \mspm{8.10}{14.03}  & \mspm{584.4}{373.6} \\
& MTA-RL (Ours)         & \bmspm{13.53}{0.85}  & \bmspm{0.764}{0.116} & \bmspm{4379.6}{516.4} & \mspm{0.233}{0.058} & \mspm{11.39}{1.76} & \bmspm{1073.8}{369.8} \\
\midrule

\multirow{4}{*}{Town 2}
& VLM-RL       & \mspmNA & \mspm{0.673}{0.169} & \mspm{1367.8}{372.4}  & \mspm{0.567}{0.231} & \mspm{13.96}{2.14} & \mspmNA \\
& Think2Drive  & \mspm{11.35}{0.87}  & \mspm{0.312}{0.055}  & \mspm{648.2}{135.8}   & \mspm{0.167}{0.058} & \mspm{23.33}{5.39} & \mspm{70.2}{18.2} \\
& Roach        & \bmspm{12.46}{1.23}  & \mspm{0.765}{0.050} & \mspm{1528.3}{143.9} & \bmspm{0.067}{0.058} & \mspm{16.13}{13.97} & \mspm{318.2}{95.2} \\
& MTA-RL (Ours)          & \mspm{11.56}{0.54}  & \bmspm{0.834}{0.118} & \bmspm{1697.1}{238.1} & \mspm{0.100}{0.100} & \bmspm{7.61}{6.59} & \bmspm{700.8}{358.2} \\
\midrule

\multirow{4}{*}{Town 3}
& VLM-RL       & \mspmNA & \mspm{0.728}{0.092}  & \mspm{2748.8}{329.5} & \mspm{0.467}{0.115} & \mspm{19.65}{2.97} & \mspmNA \\
& Think2Drive & \mspm{13.88}{0.83}  & \mspm{0.259}{0.017} & \mspm{774.3}{99.9}  & \bmspm{0.167}{0.058} & \mspm{26.81}{11.64} & \mspm{86.0}{11.1} \\
& Roach        & \mspm{12.32}{0.92}  & \mspm{0.758}{0.115} & \mspm{2858.9}{409.7} & \mspm{0.267}{0.058} & \mspm{15.18}{3.04}  & \bmspm{737.4}{339.5} \\
& MTA-RL (Ours)         & \bmspm{14.75}{0.85}  & \bmspm{0.761}{0.101} & \bmspm{2886.7}{527.5} & \mspm{0.267}{0.115} & \bmspm{14.34}{3.07} & \mspm{628.6}{157.8} \\
\bottomrule
\end{tabular}

\vspace{0.25cm}

\caption{Evaluation results in normal traffic (40 BVs). Best results are highlighted in light blue.}
\label{tab:generalization_40}
\begin{tabular}{c l c c c c c c}
\toprule
Town & Model & AS$\uparrow$ & RC$\uparrow$ & TD$\uparrow$ & CR$\downarrow$ & CS$\downarrow$ & DPV$\uparrow$ \\
\midrule

\multirow{4}{*}{Town 1}
& VLM-RL       & \mspmNA & \mspm{0.737}{0.078} & \mspm{3907.3}{620.4} & \mspm{0.567}{0.058} & \mspm{14.87}{4.87} & \mspmNA \\
& Think2Drive & \mspm{11.76}{0.56}  & \mspm{0.123}{0.009} & \mspm{698.5}{50.2}   & \mspm{0.100}{0.000} & \mspm{21.78}{7.78} & \mspm{69.8}{5.0} \\
& Roach        & \bmspm{14.99}{1.22}  & \mspm{0.572}{0.027} & \mspm{2833.5}{207.4} & \bmspm{0.067}{0.058} & \bmspm{7.13}{9.10}  & \mspm{448.7}{35.7} \\
& MTA-RL (Ours)         & \mspm{13.32}{0.32}  & \bmspm{0.866}{0.109} & \bmspm{4733.0}{770.7} & \bmspm{0.067}{0.058} & \mspm{9.51}{10.42} & \bmspm{1758.2}{888.4} \\
\midrule

\multirow{4}{*}{Town 2}
& VLM-RL       & \mspmNA & \mspm{0.760}{0.037} & \mspm{1576.8}{54.2}  & \mspm{0.433}{0.153} & \mspm{17.47}{6.94} & \mspmNA \\
& Think2Drive  & \bmspm{13.37}{0.50}  & \mspm{0.336}{0.011}  & \mspm{713.9}{20.0}   & \mspm{0.167}{0.058} & \mspm{27.65}{4.10} & \mspm{79.3}{2.2} \\
& Roach        & \mspm{11.79}{0.95}  & \mspm{0.784}{0.092} & \mspm{1594.5}{254.2} & \bmspm{0.033}{0.058} & \bmspm{6.52}{11.29} & \mspm{412.5}{193.1} \\
& MTA-RL (Ours)         & \mspm{11.95}{1.28}  & \bmspm{0.819}{0.052} & \bmspm{1721.6}{135.1} & \mspm{0.133}{0.058} & \mspm{16.44}{7.19} & \bmspm{674.8}{205.4} \\
\midrule

\multirow{4}{*}{Town 3}
& VLM-RL       & \mspmNA & \mspm{0.798}{0.061}  & \mspm{3162.3}{281.4} & \mspm{0.433}{0.115} & \mspm{14.31}{5.26} & \mspmNA \\
& Think2Drive  & \mspm{14.19}{2.97}  & \mspm{0.239}{0.073} & \mspm{658.6}{376.0}  & \bmspm{0.133}{0.115} & \mspm{22.63}{19.77} & \mspm{73.2}{41.8} \\
& Roach        & \mspm{13.10}{0.84}  & \bmspm{0.897}{0.068} & \bmspm{3513.3}{352.7} & \mspm{0.133}{0.153} & \bmspm{4.14}{7.16}  & \mspm{1083.7}{262.4} \\
& MTA-RL (Ours)          & \bmspm{15.73}{2.16}  & \mspm{0.856}{0.125} & \mspm{3394.7}{605.3} & \mspm{0.167}{0.153} & \mspm{13.51}{12.08} & \bmspm{1907.9}{1901.9} \\
\bottomrule
\end{tabular}

\vspace{0.25cm}

\caption{Evaluation results in sparse traffic (20 BVs). Best results are highlighted in light blue.}
\label{tab:generalization_20}
\begin{tabular}{c l c c c c c c}
\toprule
Town & Model & AS$\uparrow$ & RC$\uparrow$ & TD$\uparrow$ & CR$\downarrow$ & CS$\downarrow$ & DPV$\uparrow$ \\
\midrule

\multirow{4}{*}{Town 1}
& VLM-RL       & \mspmNA & \mspm{0.789}{0.087} & \mspm{4294.4}{620.6} & \mspm{0.400}{0.100} & \mspm{15.82}{3.16} & \mspmNA \\
& Think2Drive  & \mspm{12.85}{0.54}  & \mspm{0.131}{0.009} & \mspm{741.7}{38.5}   & \bmspm{0.033}{0.058} & \bmspm{5.77}{10.00} & \mspm{74.2}{3.9} \\
& Roach        & \mspm{13.74}{0.60}  & \mspm{0.638}{0.010} & \mspm{3193.0}{126.0} & \mspm{0.067}{0.058} & \mspm{12.12}{10.73}  & \mspm{602.8}{61.7} \\
& MTA-RL (Ours)          & \bmspm{14.47}{1.72}  & \bmspm{0.929}{0.060} & \bmspm{5162.2}{488.7} & \bmspm{0.033}{0.058} & \mspm{9.14}{15.83} & \bmspm{3131.6}{2280.5} \\
\midrule

\multirow{4}{*}{Town 2}
& VLM-RL       & \mspmNA & \mspm{0.804}{0.137} & \mspm{1638.8}{327.4}  & \mspm{0.333}{0.208} & \mspm{13.71}{12.6} & \mspmNA \\
& Think2Drive  & \mspm{12.97}{1.43}  & \mspm{0.339}{0.009}  & \mspm{712.8}{26.9}   & \mspm{0.167}{0.058} & \mspm{23.43}{5.30} & \mspm{79.2}{3.0} \\
& Roach        & \mspm{12.53}{1.28}  & \mspm{0.741}{0.134} & \mspm{1497.5}{291.2} & \bmspm{0.000}{0.000} & \bmspm{0.00}{0.00} & \mspm{468.1}{387.6} \\
& MTA-RL (Ours)          & \bmspm{13.39}{1.95}  & \bmspm{0.843}{0.070} & \bmspm{1727.9}{110.8} & \mspm{0.133}{0.058} & \mspm{7.59}{7.48} & \bmspm{728.8}{281.8} \\
\midrule

\multirow{4}{*}{Town 3}
& VLM-RL       & \mspmNA & \mspm{0.845}{0.093}  & \mspm{3345.9}{492.0} & \mspm{0.233}{0.153} & \mspm{15.98}{3.43} & \mspmNA \\
& Think2Drive  & \mspm{14.79}{2.30}  & \mspm{0.295}{0.014} & \mspm{951.6}{62.3}  & \bmspm{0.000}{0.000} & \bmspm{0.00}{0.00} & \mspm{105.7}{6.9} \\
& Roach        & \mspm{11.85}{1.98}  & \mspm{0.843}{0.035} & \mspm{3317.7}{350.1} & \mspm{0.133}{0.058} & \mspm{4.67}{4.60}  & \mspm{1105.9}{116.7} \\
& MTA-RL (Ours)         & \bmspm{16.03}{1.05}  & \bmspm{0.866}{0.081} & \bmspm{3439.0}{340.6} & \mspm{0.100}{0.100} & \mspm{14.40}{18.15} & \bmspm{1447.6}{568.0} \\
\bottomrule
\end{tabular}

\end{table*}

We further evaluate driving performance in complex urban environments across three CARLA towns: Town01, Town02, and Town03. Following \cite{vlm-rl}, we use 10 fixed test routes per town to ensure consistent and reproducible evaluation. To assess robustness under varying interaction complexity, three traffic density levels are considered: dense traffic with 60 background vehicles (BVs), normal traffic with 40 BVs, and sparse traffic with 20 BVs. Episodes terminate upon collision, traffic-rule violation, or if the vehicle is stuck for more than 50 steps. We compare against three strong baselines: \emph{VLM-RL}\cite{vlm-rl}, a recently proposed RL method that generates rewards based on vision language models; \emph{Think2Drive}\cite{li2024think2drive}, a world-model-based RL approach (re-implemented using the original hyperparameters since official code is unavailable); and \emph{Roach}\cite{zhang2021roach}, a classic autonomous driving planner trained using the PPO algorithm.

Performance is evaluated using Average Speed (\textbf{AS}); Route Completion (\textbf{RC}), the ratio of traveled to total route distance; Total Distance (\textbf{TD}) driven; Collision Rate (\textbf{CR}); Collision Speed (\textbf{CS}) at impact; and Distance Per Violation (\textbf{DPV}), the mean distance between consecutive infractions. Since VLM-RL does not explicitly model traffic rules or pedestrians, AS and DPV are omitted for this baseline. All test routes are evaluated three times in each town, and the mean and standard deviation of each metric are reported in Tables~\ref{tab:generalization_60}–\ref{tab:generalization_20}.\\

Under dense traffic conditions with 60 BVs, Table~\ref{tab:generalization_60} shows that the proposed method consistently achieves the highest RC and TD across all three towns, indicating strong generalization to previously unseen and highly interactive urban scenarios. In terms of driving efficiency, our method attains the highest AS in Town01 and Town03, while in Town02 its average speed is comparable to \emph{Roach}, with a difference of only 0.9~km/h. Regarding safety, although \emph{Think2Drive} yields the lowest CR in Town01 and Town03, its frequent violations in other aspects lead to substantially lower RC and TD. In contrast, our method achieves the lowest CS in Town02 and Town03 and the largest DPV in Town01 and Town02, indicating fewer violations and safer overall behavior. Notably, even though \emph{VLM-RL} operates in a simplified setting without explicit traffic rules or pedestrian interactions, the proposed method still outperforms it across all reported metrics. As traffic density decreases, all methods improve. Nevertheless, the proposed framework consistently achieves the best or near-best RC, TD and DPV across towns, maintaining strong safety-efficiency trade-offs and stable cross-town generalization.

\subsection{Ablation Study}

To evaluate the contribution of each component in the proposed \emph{MTA-RL} framework, we conduct a series of ablation studies.

\subsubsection{Multi-modal 3D Affordance}

We investigate two key design choices: (1) multi-modal fusion of RGB and LiDAR inputs, and (2) transformer-based cross-modal feature integration. Following \cite{chitta2022transfuser}, we implement two baselines. \textit{Latent TransFuser} replaces the LiDAR BEV histogram with a 2-channel positional encoding of identical size, removing explicit geometric information. \textit{Geometric Fusion} replaces the transformer module with a multi-scale geometry-based fusion strategy inspired by \cite{liang2018deep}. Results in Table~\ref{tab:ablation affordance} show that combining RGB and LiDAR with transformer-based fusion yields the best performance, validating the importance of explicit geometric input and attention-based multi-modal reasoning for accurate affordance prediction.

\begin{table}
    \centering
    \caption{Affordance prediction results on the test set for ablation study. Best performance for each task is highlighted in bold.}
    \label{tab:ablation affordance}
    \small
    \setlength{\tabcolsep}{4pt}
    \renewcommand{\arraystretch}{1.1}
    \resizebox{0.95\columnwidth}{!}{%
    \begin{tabular}{lccc}
    \toprule
    \textbf{Affordance} & \textbf{Latent TransFuser} & \textbf{Geometric Fusion} & \textbf{MTA-RL(Ours)} \\
    \midrule
    \addlinespace[0.25em]
    \multicolumn{4}{@{}l@{}}{\textbf{Categorical} (IoU $\uparrow$)}\\
    \addlinespace[0.15em]
    Red traffic light & \textbf{99.41\%} & 97.64\% & 99.13\%\\
    Stop sign & 91.68\% & 87.44\% & \textbf{92.25\%}\\
    Hazard stop  & 97.63\% & 87.34\% & \textbf{98.76\%}\\
    \addlinespace[0.15em]
    \midrule
    \addlinespace[0.2em]
    \multicolumn{4}{@{}l@{}}{\textbf{Continuous} (MAE $\downarrow$)}\\
    \addlinespace[0.1em]
    Lateral distance  & 0.185 & 0.199 & \textbf{0.124}\\
    Relative angle  & \textbf{0.012} & 0.124 & 0.019\\
    Target vehicle distance  & 0.154 & 0.054 & \textbf{0.011}\\
    Traffic light distance  & 0.542 & 0.424 & \textbf{0.312}\\
    Stop sign distance  & 0.446 & 0.438 & \textbf{0.264}\\
    Pedestrian distance & 0.314 & 0.476 & \textbf{0.236}\\
    \bottomrule
    \end{tabular}%
    }
\end{table}

\subsubsection{Dense Reward Shaping}

\begin{figure}
    \centering
    \includegraphics[width=\columnwidth]{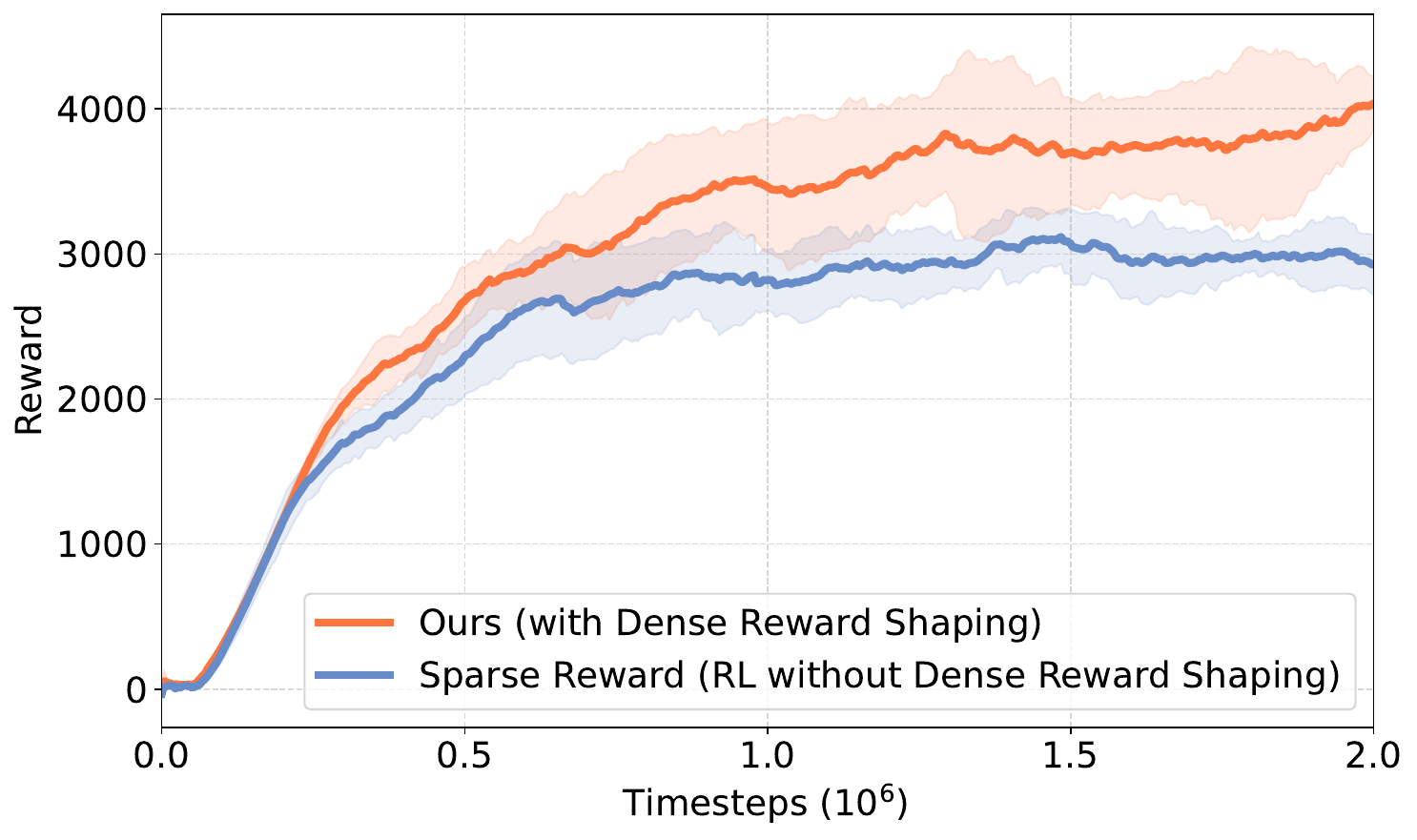}
    \caption{Reward curves for the dense reward shaping ablation.}
    \label{fig:reward_curve}
\end{figure}

For the RL controller, we ablate the dense traffic-rule shaping term $R_{\text{rule}}(t)$ by training a variant that relies solely on sparse terminal penalties (Table~\ref{tab:terminal_penalty}). As shown in Fig.~\ref{fig:reward_curve}, dense shaping leads to faster convergence and consistently higher cumulative returns. Table \ref{tab:ablation_re} further shows that, while average speeds are similar, dense shaping consistently improves RC, TD, and DPV across all test towns. This improvement stems from more informative intermediate feedback, which alleviates long-horizon credit assignment issues and reduces inefficient exploration. In contrast, sparse rewards provide delayed and coarse signals, limiting policy refinement in complex urban scenarios. These results demonstrate that dense traffic rule shaping stabilizes learning and enhances safe, robust decision making.

\begin{table}
\centering
\footnotesize
\setlength{\tabcolsep}{3pt}
\renewcommand{\arraystretch}{1.05}
\caption{Evaluation results of the reward shaping ablation study.}
\label{tab:ablation_re}
\resizebox{\columnwidth}{!}{
\begin{tabular}{c l c c c c c c}
\toprule
Town & Model & AS$\uparrow$ & RC$\uparrow$ & TD$\uparrow$ & DPV$\uparrow$ \\
\midrule
\multirow{2}{*}{Town 1}
& Sparse Reward       & \mspm{13.03}{1.39} & \mspm{0.727}{0.123} & \mspm{4007.2}{504.1}  & \mspm{906.4}{290.1} \\
& MTA-RL (Ours)         & \bmspm{13.53}{0.85}  & \bmspm{0.764}{0.116} & \bmspm{4379.6}{516.4} & \bmspm{1073.8}{369.8} \\
\midrule

\multirow{2}{*}{Town 2}
& Sparse Reward        & \bmspm{11.57}{1.56} & \mspm{0.744}{0.104} & \mspm{1490.8}{286.8}  & \mspm{314.2}{125.2} \\
& MTA-RL (Ours)          & \mspm{11.56}{0.54}  & \bmspm{0.834}{0.118} & \bmspm{1697.1}{238.1} & \bmspm{700.8}{358.2} \\
\midrule

\multirow{2}{*}{Town 3}
& Sparse Reward        & \bmspm{15.22}{0.86} & \mspm{0.740}{0.061}  & \mspm{2730.0}{297.2} & \mspm{596.0}{138.5} \\
& MTA-RL (Ours)         & \mspm{14.75}{0.85}  & \bmspm{0.761}{0.101} & \bmspm{2886.7}{527.5} & \bmspm{628.6}{157.8} \\
\bottomrule
\end{tabular}
}
\end{table}

\section{CONCLUSION}

We present \emph{MTA-RL}, a robust urban driving framework that tightly integrates multi-modal transformer-based perception with RL through structured 3D affordances. By fusing RGB and LiDAR inputs using attention-based cross-modal interactions, our perception module produces compact, interpretable affordances that explicitly encode both semantic traffic constraints and geometric relationships in the environment. Operating the RL policy in this affordance space significantly improves sample efficiency, stability, and generalization compared to policies trained on raw sensory inputs or unstructured state representations. Ablation studies validate the contribution of each component in the proposed framework. Extensive evaluations in CARLA across multiple towns and traffic densities demonstrate that \emph{MTA-RL} consistently outperforms state-of-the-art RL baselines. Overall, this work highlights the effectiveness of multi-modal 3D affordances as an intermediate representation, providing a principled balance between end-to-end learning and modular design for urban autonomous driving.

\bibliographystyle{IEEEtran}
\bibliography{refs}

\end{document}